\documentclass[11pt]{article}

\usepackage[preprint]{acl}

\usepackage{times}
\usepackage{latexsym}

\usepackage[T1]{fontenc}

\usepackage[utf8]{inputenc}

\usepackage{microtype}

\usepackage{inconsolata}

\usepackage{graphicx}

\usepackage{booktabs}
\usepackage{fancyvrb}
\usepackage{multicol}
\usepackage{hyperref}
\usepackage{amsmath}

\usepackage{enumitem}

\usepackage{cjhebrew}
\newcommand{\EN}[1]{\beginL\textnormal{#1}\endL}
\newenvironment{hpar}{\par\begin{cjhebrew}}{\end{cjhebrew}\par}

\usepackage{xcolor}
\usepackage{mdframed}
\usepackage[T1]{fontenc}          
\usepackage[most]{tcolorbox}
\usepackage{enumitem}
\usepackage{fvextra}              

\title{From DQI to C-DQI: Automatic Codebook Generation and Evaluation \\ for Measuring Computational Discourse Quality Index}

\author{
  Esther Shizgal$^{1*}$ \quad Ron Keinan$^{1*}$ \quad Daniel Shtauber$^2$ \\[0.5ex] 
  \textbf{Roee Tzvi Esquira$^1$ \quad Matan Toledano$^1$ \quad Or Rappel-Kroyzer$^3$ \quad Liron Lavi$^4$} \\[0.2ex]
  \textbf{Ori Aronson$^5$ \quad Renana Keydar$^{3,6}$ \quad Ittai Bar-Siman-Tov$^5$ \quad Reut Tsarfaty$^1$} \\[0.8ex] 
  $^1$Department of Computer Science, Bar-Ilan University \\
  $^2$Max Planck Center for Democracy, Security and Human Rights, Faculty of Law, Hebrew University \\
  $^3$Digital Humanities, Hebrew University \\
  $^4$Department of Political Studies, Bar-Ilan University \\
  $^5$Faculty of Law, Bar-Ilan University \\
  $^6$Faculty of Law, Hebrew University \\[0.6ex] 
  $^{1,4,5}$\texttt{\{first\_name\}.\{last\_name\}@biu.ac.il} \\ 
  $^{2,3,6}$\texttt{\{first\_name\}.\{last\_name\}@mail.huji.ac.il}
}

\begin{document}
\maketitle
\begingroup\def\thefootnote{*}\footnotetext{Equal contribution.}\endgroup
\begin{abstract}

Effective democratic deliberation is vital for civic engagement and the health of legislative institutions. However, assessing parliamentary discourse quality is traditionally done using expert-scored, theory-based {\em codebooks} (CBs), which are rigid, labor-intensive, and prone to biases. While Large Language Models (LLMs) offer a scalable alternative for both schema induction and discourse annotation, ensuring schema quality and mitigating models' inherent bias remains a clear challenge. In this work, we propose a complete protocol for LLM-based codebook generation and evaluation. Starting from the widely accepted {\em discourse quality Index} (DQI), we compare top-down (theory-based), bottom-up (data-driven), and hybrid approaches for generating context-adapted schemas. To evaluate these schemas, we define an extrinsic proxy measure, validated by manual annotations, alongside an intrinsic {\em inter-model agreement} (IMA) metric, which leverages differing model biases to signal overall reliability. Our experiments on 100 Israeli parliament transcripts reveal that inductive, data-driven schemas achieve higher IMA than top-down methods while maintaining validity, demonstrating their potential for reliable, large-scale analysis of political discourse.
\end{abstract}

\begin{figure*}[t]
\centering
\includegraphics[width=0.95\linewidth]{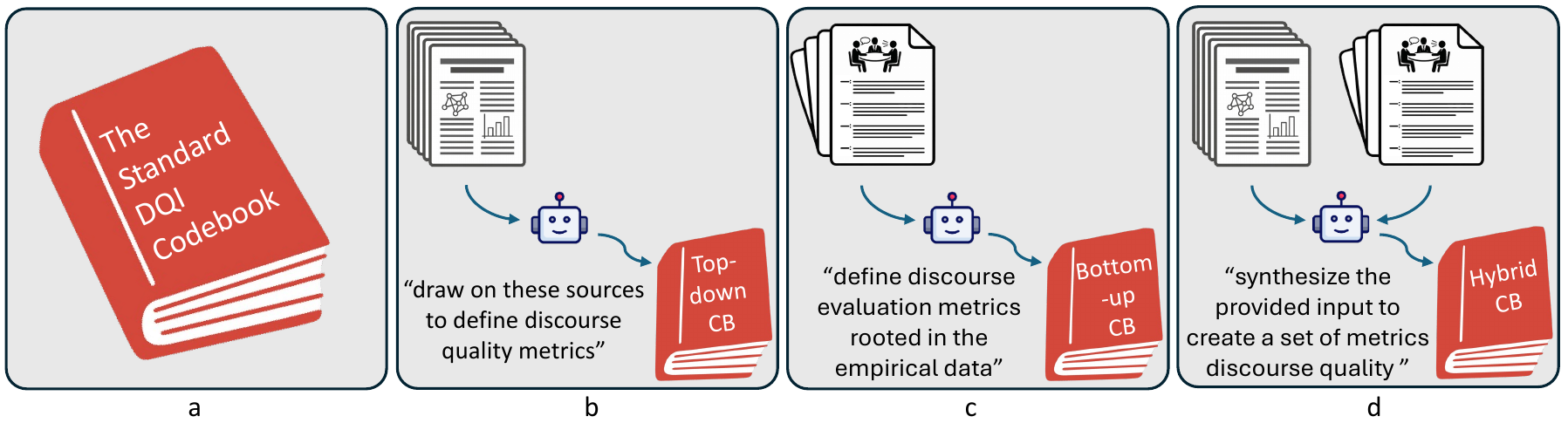}
\caption{The four workflows for deriving CBs for deliberation quality assessment: (a) the standard DQI CB (baseline), and three LLM-generated CBs ---  (b) theory-led (top-down), (c) data-driven (bottom-up), and (d) a hybrid CB combining the two previous approaches.}
\label{workflows}
\end{figure*}

\begin{figure*}[t]
\centering
\includegraphics[width=0.95\linewidth]{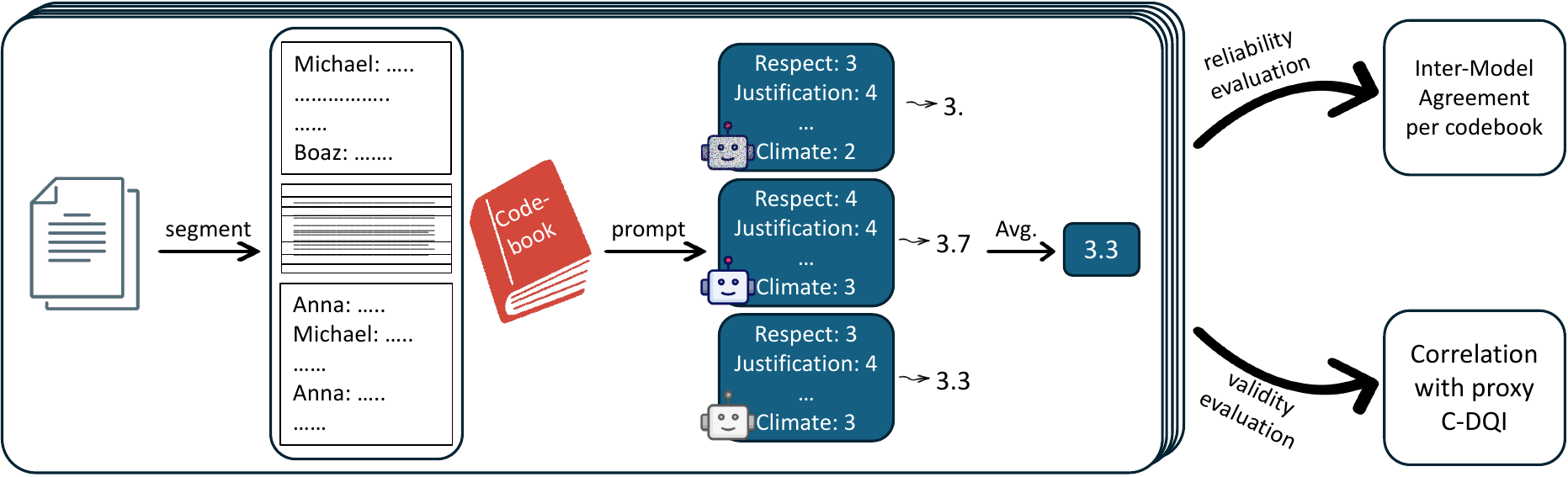}
\caption{The evaluation pipeline: each discussion is segmented and scored against a given CB by multiple models, whose outputs are averaged. Repeating this across discussions yields the CB's reliability (inter-model agreement) and validity (rank correlation with proxy C-DQI).}
\label{pipeline_ima}
\end{figure*} 

\section{Introduction}
Deliberation is the interactive process where individuals engage in discussion to reach rational conclusions, occurring in both formal political bodies and local civic settings. At its core lies the premise that political decision-making should be \textit{talk-centric} rather than \textit{vote-centric} \citep{steenbergen2003measuring}. Effective democratic deliberation is crucial for civic engagement and the health of legislative institutions \citep{serranocontreras2020measuring, schroeder2024, tschentscher2010deliberation}. Prominent literature identifies thoughtful analysis, open-mindedness, mutual respect, and participants' willingness to revise their initial preferences through discourse as prerequisites for this efficacy \citep{steenbergen2003measuring, ercan2022}. 

Quantifying deliberation quality enables its application to downstream tasks, such as identifying anomalous deliberative behavior, evaluating the performance of legislators, and alerting civic facilitators when participant satisfaction could be improved \citep{schroeder2024}. But while numerous factors have been argued to affect discourse quality, such as domination, issue polarization, and agenda setting \citep{kumar2025advancing, serranocontreras2020measuring}, quantifying this quality remains challenging due to the inherent competing interests between interlocutors, rhetorical ambiguity, and the variance in acceptable standards across societies and domains.

Various methods, including computational frameworks, have been proposed for measuring  multi-party discourse quality \citep{steenbergen2003measuring, behrendt2024aqua, bosley2025towards}.  A common protocol for quality assessment involves experts evaluating live multi-party discourse using a theoretical-scientific scoring schema. This schema, known as a \textit{codebook} (CB), comprises categories that reflect rationality, reciprocity, and civility, and uses ordinal scales to individually assess each criterion \citep{behrendt2024aqua, graham2010, esau2021, bachtiger2010, steenbergen2003measuring}.
However, this paradigm has various limitations. First, rigid, theory-based schemas often struggle to generalize across topic-dependent or culture-dependent nuances. Second, collecting human annotation is labor-intensive and inherently subject to individual biases \citep{romberg2025perspectivist, vecchi2025perspectivemod, brown2025evaluating}. 

To automate the application of such schemas, researchers have employed supervised machine learning \cite{fournier2020delibanalysis, fournier2021big, Berkovitz2023open} and Large Language Models (LLMs) \citep{bosley2025towards, langlais2025legibot, chen2026large}. However, both approaches still rely upon the same rigid frameworks.

In this work, we propose {\em Computational Discourse Quality Index} (C-DQI), a methodology for the definition and application of CBs for assessing discourse quality.  We explore four distinct workflows for C-DQI CB generation. We start with a baseline adapted from \citet{steenbergen2003measuring}'s Discourse Quality Index (DQI) and implement three pipelines that deploy LLMs for CB induction, differing in their balance of top-down (theory-based) and bottom-up (data-driven) approaches.

To evaluate the CBs, we propose both intrinsic and extrinsic measures. Following \citet{steenbergen2003measuring}, who discuss the importance of robustness in a discourse quality index,  and acknowledging that LLMs are inherently pre-trained with internal biases \citep{mor2026location}, we argue that the consensus of multiple models indicates that the CBs’ criteria are reliable for LLM annotation, rather than artifacts of a single model. We thus propose to define  {\em Inter-Model Agreement} (IMA) scores across multiple LLMs to capture intrinsic reliability.
A higher IMA score indicates a clearer, more objective, and robust framework for automated protocol annotation. However, because agreement does not equate to validity, we further compare the CBs' overall assessments against human-validated proxies for deliberation quality, utilizing both manual and automatic annotations. \autoref{workflows} demonstrates the full C-DQI pipeline and evaluation process.

To assess the efficacy of our proposed workflows, we apply different LLMs to annotate a corpus of 100 committee meeting transcripts from the Israeli parliament, the Knesset, according to each CB. The data spans multiple governments and a diverse range of bills, providing a dynamic, culture-dependent environment for examining discourse quality. Utilizing this region-specific data demonstrates the cultural alignment challenges inherent in analyzing localized political discourse.

Our results reveal that bottom-up CBs exhibit higher IMA, while all CBs correlate comparably with the quality proxy ($\rho$ > 0.58), demonstrating that data-driven schema generation yields more reliable evaluations without sacrificing validity.
Through this work, our primary contributions are twofold: (1) we develop a computational tool that automatically defines and evaluates discourse quality CBs for multi-party discussions and scores the data accordingly; and (2) we provide empirical insights into the optimal balance between top-down theoretical frameworks and bottom-up, inductive approaches for yielding reliable evaluation criteria.

\section{Background}
\paragraph{Discourse Quality Measures. }
A widely applied measure for deliberative quality is the Discourse Quality Index (DQI) originally developed for the analysis of parliamentary debates \citep{steenbergen2003measuring}. 
The DQI is theoretically grounded in established discourse ethics and its schema includes criteria representing norms that discourse is expected to satisfy. Its unit of analysis is an individual speech, defined as ``the public discourse by a particular individual delivered at a particular point in a debate'' (\S\ref{sec:vanilla_dqi} provides further details).

Yet, this de facto standard is constrained by drawbacks tied directly to the DQI framework, as well as broader limitations inherent to predefined schemas. First, despite well-defined coding categories, the DQI does not prescribe a standardized procedure for aggregating category-level annotations into a single discourse-level score, complicating its use in large-scale automated evaluation \citep{bachtiger2005appendix}. Second, according to the DQI, off-topic discourse and clarifying questions are excluded from the final score, and interruptions are only accounted for if participants complain about them, ignoring the potential for these factors to contribute to the overall quality assessment.

More generally, hard-coded CBs target specific settings and contexts, and therefore struggle to generalize to topic-dependent or culture-dependent contexts; and manually annotating them is labor-intensive and inherently subject to individuals' biases \citep{romberg2025perspectivist, vecchi2025perspectivemod, brown2025evaluating}.

\paragraph{Computational Discourse Quality Evaluation.}
Multiple efforts have operationalized discourse quality assessment by applying supervised machine learning to the DQI \citep{fournier2020delibanalysis, fournier2021big, Berkovitz2023open}. Furthermore, \citet{kumar2025advancing} recently combined argumentation graphs with the DQI for more objective assessment. Deep learning approaches like the AQuA framework successfully bridge NLP with deliberation theory to analyze discourse semantics, though they remain resource-intensive due to the overhead of running multiple fine-tuned BERT adapters \citep{behrendt2024aqua}.

Currently, the field is actively moving toward using LLMs as evaluative agents, bypassing the limitations of manual annotation and rigid linguistic features \citep{langlais2025legibot, chen2026large, bosley2025towards}. Specifically, \citet{bosley2025towards} evaluated the DQI using a few-shot prompting framework, achieving IMA comparable to human standards.  Their work laid the foundation for our methodological study on C-DQI schemas.

\paragraph{Automatic Schema Discovery.}
To address the limitations of rigid, predefined schemas, researchers have utilized LLMs for automatic schema definition. Earlier work deploys LLMs to construct literature-review tables by prompting the model to generate a schema and then populating the table from the underlying data \citep{newman2024arxivdigestables}, optionally conditioned on a user-provided intent \citep{padmakumar2025intent}. 

More recently, human-in-the-loop systems have been developed to automate schema definition over large corpora of unstructured text, including ScheMatiQ \citep{levy2026schematiq} and Schema-miner\textsuperscript{pro} \citep{sadruddin2025llms4schemadiscovery, sadruddin2026schemaminerpro}. These systems combine prompting pipelines with interactive interfaces and agentic tooling. 

To our knowledge, however, the outputs of these systems have been evaluated mainly for their agreement with expert-annotated schemas, rather than being systematically assessed for intrinsic reliability, which we emphasize in our current work.

\section{The Challenge}
Without systematically evaluating \textit{codebook} efficacy, the question remains:  How can we derive new or adapted schemas for new settings and domains?  How can we determine whether a schema is {\em good}? And, how can we avoid the undermining effects of inherent biases of individual LLMs?

\section{A  Complete Protocol for LLM-based Codebook Generation and Evaluation}

To remove the reliance on rigid CBs that require manual annotation,
we propose an LLM-based methodology for effective and efficient definition and application of a {\em Computational Discourse Quality Index} (C-DQI). This methodology addresses the core limitations of traditional schema-based annotations by allowing for flexible schema induction across different contexts and replacing human annotators with LLMs to mitigate subjective human biases.

In \S\ref{sec:cdqi} we introduce C-DQI, used to quantify quality based on a codebook. In \S\ref{sec:cb_gen} we then define the computation task of codebook induction and its various variants, and finally in  \S\ref{sec:cb_eval} we define metrics that aim to assess the reliability and validity of the generated codebook.

\subsection{Computational Discourse Quality Index}
\label{sec:cdqi}

We next define the notion of a {\em Computational Discourse Quality Index} (C-DQI) which is our means to define discourse quality in quantitative terms. 
Let $D$ be an input transcript and $C$ be a given CB.   The input transcript $D$ is split into multiple segments $D = (s_1, s_2, \dots, s_n)$, and the CB $C$ defines a set of evaluation categories $C = \{c_1, c_2, \dots, c_m\}$ where each category is scored on a predefined scale.\footnote{We use a 4-point scale. Because DQI assigns each category its own scale, we min-max normalize its raw scores to this range: $\tilde{y} = 1 + 3 \frac{y - y_{\min}}{y_{\max} - y_{\min}}$, where $y_{\min}, y_{\max}$ are the category's minimum and maximum possible scores. This keeps all CBs comparable.}
For every segment-category pair $(s_i, c_j)$, an LLM annotator $M$ evaluates the text on the required scale, yielding a score $y_{i,j}$. The per-category discussion score, $S(c_j)$, is calculated as the average across all segments:
$$S(c_j) = \frac{1}{n} \sum_{i=1}^{n} y_{i,j}$$
This final index, representing the overall discourse quality for the transcript, is the average over all categories:
$$\text{C-DQI}(D, M) = \frac{1}{m} \sum_{j=1}^{m} S(c_j)$$
When a single comprehensive score per transcript is required, we report the mean of the C-DQI scores produced by the ensemble of annotator models.

\subsection{LLM-based Codebook Generation}
\label{sec:cb_gen}

We define the task of CB generation as a function from texts to a set of categories. Formally, let $T$ be a collection of texts and let $C=\{c_1, \dots, c_n\}$ be a finite set of categories describing dimensions for evaluating discourse; the CB generation task is defined as $f(T) = C$.

Specifically, we implement four workflows to derive CBs.
One is a standard baseline, and three are LLM-based --- one receives academic literature as the input text ({\em top-down}), the second reviews transcripts from the domain being studied ({\em bottom-up}), and the third combines both inputs ({\em hybrid}) (see \autoref{workflows}). All prompts are available in \autoref{appendix_cb_prompts}.

\paragraph{Vanilla C-DQI.}
\label{sec:vanilla_dqi}

The baseline workflow adapts the original Discourse Quality Index (DQI), which features seven coding categories: (1) participation, (2) level of justification, (3) content of justification, (4) respect for the groups to be helped by particular policies, (5) respect toward the demands of others, (6) respect toward counterarguments, and (7) constructive politics. 

Deriving the final score involves coding all speeches and interpreting their criteria scores. The authors define relevant speeches as ones containing a policy proposal (originally termed a `demand'). 

The coding process requires initial filtering to distinguish relevant excerpts, which are then scored per category. For this workflow, we adapt the authors' original coding guidelines \citep{bachtiger2005appendix} by assigning numerical values to their scales and adjust the unit of analysis from individual speeches to segments of up to 1,500 words to maintain consistency with our other CBs.

\paragraph{Top-Down C-DQI.}
\label{sec:lit}

To define the theory-driven ({\em Top-Down}) CB, we provide an LLM with a selection of academic literature on deliberation to generate a set of categories. We begin by prompting the literature-survey tool developed by \citet{literature-survey-skill} with the instruction: ``List 10 foundational academic papers on democratic deliberation, in order to develop a discourse quality index'', followed by an iterative process of refining the results according to our prior knowledge and available sources.

The resulting list ranges from sources studying Habermas's discourse ethics \citep{heath2019discourse} to methodological articles \citep{steenbergen2003measuring, stromergalley2007measuring}. The selection also includes frameworks for evaluating deliberative quality \citep{mansbridge2012systemic} and sources providing dimensions for evaluation rooted in deliberative theory \citep{friess2015systematic}.

We then instruct Gemini-3.1-Pro\footnote{\url{https://gemini.google.com/}} to define metrics for evaluating the quality of discourse in the selected dataset, based on the frameworks from the selected sources. The response serves as the Top-Down CB.

\paragraph{Bottom-Up C-DQI.}
\label{sec:bottom-up}

To define the inductive ({\em Bottom-Up}) CB, we prompt Gemini-3.1-Pro with 10 random transcripts to generate evaluation metrics from empirical data. To prevent contamination, these transcripts are excluded from later experiments. Ultimately, this workflow yields three CB variants: {\em Bottom-Up-English}, generated from an English prompt; {\em Bottom-Up-Hebrew}, generated from a Hebrew prompt; and {\em Bottom-Up-Corrected}, which we, as native Hebrew speakers, merged from the two, edited for content, and standardized in structure (\autoref{cb_corrections}).

\paragraph{Hybrid C-DQI.}
The Hybrid C-DQI CB combines both the theory-led and inductive approaches. We generate this CB by instructing Google's Gemini Notebook \footnote{\url{https://notebooklm.google.com}} to synthesize the previously selected academic sources with the empirical transcripts, to create a unified set of metrics for evaluating discourse quality in general committee meetings.

\subsection{LLM-based Codebook Evaluation}
\label{sec:cb_eval}

\paragraph{CB Reliability.}

We define the measure of reliability as the consistency of the generated CBs, calculated based on IMA scores and averaged over each CB's criteria. 

\citet{steenbergen2003measuring} name four properties that a measure of discourse quality must exhibit:  {\em theoretical grounding}, reliance on {\em observable phenomena}, {\em generalizability}, and {\em reliability}. To evaluate {\em reliability}, they use the ratio of agreement among coders, Cohen’s $\kappa$, Spearman's $\rho$, and the standardized $\alpha$.

Inter-annotator agreement serves as a lens into the complexity of the task and the quality of the annotations. Higher values indicate that the definitions are clear for different coders, whereas lower values suggest both annotator performance issues and poorly defined categories. 

We introduce Inter-Model Agreement (IMA) to quantify consensus across different models. Because these LLMs are developed by diverse organizations on distinct training data \citep{Shmidman2025DictaLM3, anthropic2026sonnet55, gemmateam2026gemma4, jiang20236g, bai2023qwen}, we hypothesize they exhibit divergent inherent biases; prior research on cultural biases validates this premise \citep{mor2026location}. This diversity contrasts with human annotators, who often share demographic backgrounds or subjective opinions. Consequently, high IMA serves as a compelling measure of robustness. If multiple models agree on the scoring criteria, it indicates that the CBs are reliable for LLM annotation, rather than reflecting the artifacts of a singular model. Thus, a higher agreement score indicates a clearer, more objective, and robust framework for automated protocol annotation. 

Following standard practices for evaluating the DQI, we report the percent agreement as a baseline measure of consensus. To rigorously account for chance agreement, we include Krippendorff's $\alpha$ \citep{krippendorff2011computing}; values range from -1 to 1; negative values indicate systematic disagreement, while values above zero indicate a degree of agreement. Furthermore, we use Spearman's $\rho$ to capture annotator agreement on the relative ranking of discussion quality, and the Standardized $\alpha$ to evaluate the overall internal consistency of the aggregated index.

\paragraph{CB Validity. }

To validate the CBs' overall performance, we compare their final scores against proxies for discourse quality. For each discussion transcript, we calculate this proxy score by prompting an independent LLM to holistically assess sequences of turns, using a short-context sliding window. These LLM assessments are validated by human annotators to ensure they adequately capture conversational dynamics. This procedure serves as an extrinsic measure of how accurately each schema captures overall discourse quality.
We define this automated proxy for every transcript, represented as a sequence of turns $D = (t_1, \dots, t_k)$. The task scores the conversation dynamics of an $n$-turn exchange on a binary scale, where $0$ reflects negative deliberation and $1$ indicates positive deliberation (see \autoref{sec:proxy_prompt} for the detailed instructions). We use a sliding window of size $n=3$ and score each window independently. The final proxy score for a conversation is the average of all its triplet scores ($p_i$):

$$P(D) = \frac{1}{k-2} \sum_{i=1}^{k-2} p_i$$

\begin{table}[t]
\centering
\footnotesize
\setlength{\tabcolsep}{3pt}
\begin{tabular}{@{}lrrrrr@{}}
\toprule
\textbf{Statistic} & \textbf{Mean} & \textbf{Median} & \textbf{Std.} & \textbf{Min} & \textbf{Max} \\
\midrule
Transcript length     & 9,393.7 & 8,414.5 & 6,545.9 & 170  & 27,113 \\
Turns / transcript    & 365.4   & 267.5   & 337.8   & 5    & 1,736  \\
Turn length          & 25.7    & 10.0    & 53.5    & 1    & 1,364  \\
Speakers per transcript & 16.9    & 17.0    & 7.9     & 2    & 39     \\
Segments per transcript & 7.0     & 6.0     & 4.5     & 1    & 19     \\
Segment length        & 1,342.0 & 1,466.0 & 299.6   & 45   & 1,500  \\
\bottomrule
\end{tabular}
\caption{Structural statistics of the 100-transcript meeting corpus. Length statistics are reported in words.}\label{tab:corpus_stats}
\end{table}

\section{Data}
\label{sec:data}

The data for our empirical assessment consists of over 15,000 hours of Hebrew recordings and corresponding transcripts. These transcripts are derived from more than 30,000 human-generated committee meeting protocols of the Knesset.\footnote{\url{https://m.knesset.gov.il/apps/committees/lobby/protocols}} This corpus spans from November 2018 to January 2026, encompassing six consecutive Knesset terms. The data was curated from government-released material and made available as a Hugging Face dataset by Ivrit.ai \citep{marmor2025building}.\footnote{\url{https://huggingface.co/datasets/ivrit-ai/knesset-committees}}
For this study, we restrict the corpus to legislative sessions by retaining only protocols in which every agenda item listed in the session metadata is marked as a bill.\footnote{Following the authors' recommendation, we select only protocols with transcription quality $>0.7$. A stratified sample of 110 (seed=20240731) provides 10 protocols for CB development and 100 for evaluation.} 

We randomly sample 100 protocols from this subset, stratified across Knesset terms and committees. \autoref{tab:corpus_stats} describes the selection's statistics.
To demonstrate the corpus' diversity
we highlight here items with notable variance:
\begin{itemize}[leftmargin=*,nosep,itemsep=3pt]
    \item[]
    \textbf{Most Intensive Meeting:} The longest session regarded the implementation of the Economic Policy for 2023 and 2024. It lasted nearly 3 hours, with 27 participants speaking, generating 1,736 turns and 27,113 words ({session\_id=2204258}).    
    \item[] 
    \textbf{Most Chair-Dominated Session:} During the Economic Affairs Committee's discussion on the bill to Increase Competition in the Broadcasting Market, the chair alone spoke 60.5\% of all words ({session\_id=2220120}).   
    \item[]
    \textbf{Most Evenly Distributed:} The most balanced discussion occurred during the National-Civic Service Bill, yielding a Gini coefficient\footnote{The Gini coefficient is a statistical measure of inequality. We apply it here to the number of words per speaker. It measures inequality on a scale from 0 to 1. A value of 0 indicates perfect equality, where every speaker uttered the same number of words. A value of 1 indicates perfect inequality, where a single person spoke for the entire meeting.} of 0.20 among five speakers and eight participants ({session\_id=2209191}). Furthermore, the most broadly inclusive \textit{large} discussion focused on the bill for reducing the use of disposable shopping bags, which featured 24 speakers, a Gini score of 0.40, and no single speaker accounting for more than 15.8\% of the words ({session\_id=2193138}). 
\end{itemize}

\vspace{5pt}
This analysis confirms our selection's potential to examine discourse quality across diverse axes of length, dominance, and inclusivity.

\section{Experimental Setup}
\paragraph{Goal. }
We aim to empirically assess the proposed LLM-based CB generation pipeline using our proposed metrics. 
We evaluate each of the four CB generation strategies, comparing the top-down, bottom-up, and hybrid scenarios.

\paragraph{Data. }
Drawing on the 100 transcript selection (\S\ref{sec:data}), 
we segment the discussion into smaller units of analysis, defined as text excerpts of up to 1,500 words and ensuring that individual turns are not split. This text length provides sufficient context for analyzing interaction dynamics and responsiveness, as the median turn length is 10 words. This results in a total of 700 segments for the full corpus, with the majority exceeding 1,466 words (\autoref{tab:corpus_stats}).

\paragraph{Models. }
Our selection includes a proprietary LLM (Claude Sonnet 5.5\footnote{{\href{https://platform.claude.com/docs/en/models/sonnet-5-5/overview}{https://platform.claude.com/docs/en/models/sonnet-5-5/overview}}}), and open-weight models: a Hebrew-centric model (DictaLM3-24B-Thinking \citep{Shmidman2025DictaLM3}), and multilingual models (Gemma4-31B \citep{gemmateam2026gemma4}, Qwen3.6-27B \citep{qwen3.6-27b}, GPT-OSS-20B \citep{openai2025gptoss120bgptoss20bmodel}, and Mistral-Small-3.2-24B\footnote{\href{huggingface.co/mistralai/Mistral-Small-3.2-24B-Instruct-2506}{mistralai/Mistral-Small-3.2-24B-Instruct-2506}}).

For the open-weight models, we use default hyperparameters with temperature 0.6. We run each model using two seeds (11, 22) and report the mean. For Claude, due to cost considerations, we perform a single run with the temperature 1.0.

For estimating the proxy C-DQI, we utilize GPT-6-luna\footnote{\href{https://developers.openai.com/api/docs/models/gpt-6-luna}{gpt-6-luna}} with its default hyperparameters.

\paragraph{CB Application. }
We prompt each model with every category-segment pair. Apart from the Vanilla C-DQI CB, which utilizes category-specific scales, all categories are scored on a four-point scale. To create a final, representative vector of deliberation criteria for an entire discussion, the segment-level scores for each category are averaged. The final C-DQI score for a given discussion is determined according to the definitions in \S\ref{sec:cdqi}.

\paragraph{Proxy Validation.}
With a team of domain experts, we manually annotated 162 exchanges to serve as a human benchmark, attaining a Krippendorff’s $\alpha$ of 0.79 for inter-annotator agreement in a complete overlap setting where each exchange was reviewed by exactly two annotators. Exchanges on which the annotators disagreed were discarded.

We then deployed GPT-6-luna\footnote{run through the OpenRouter API on October 12, 2026: \url{https://openrouter.ai/openai/gpt-6-luna-20260922}} --- an independent LLM not utilized in the CBs' reliability evaluation --- to compute proxy scores across the dataset. Against the human benchmark, this LLM achieved an accuracy of 82\% and a macro F1 score of 77\% on a held-out test set of 50 samples (see \autoref{sec:appendix_judge_selection}). Deploying an LLM as a judge not only allows us to scale the validation across the entire dataset (a total of 36,341 3-turn exchanges), avoiding costly time and resource constraints, but it also provides a consistent scoring mechanism that mitigates the noise and subjectivity inherent in human annotations. Consequently, we rely on these automated proxy values to validate our experiments. The distribution of the proxy scores across the corpus is provided in \autoref{sec:appendix_proxy_dist}.

\paragraph{Evaluation Metrics. } 

To assess the generated CBs, we employ the metrics defined in \S\ref{sec:cb_eval}. For CB reliability, we compute IMA scores for each category.\footnote{Since the DQI CB includes filtering ``relevant speeches'' (\S\ref{sec:vanilla_dqi}), for the Vanilla DQI we also compute the IMA score for this phase.} 
For validity, we calculate Spearman correlation coefficients between the mean C-DQI scores and the proxy scores.

\section{Results and Analysis}

\begin{table*}[t]
\centering
\begin{tabular}{@{}lcccc@{}}
\toprule
\textit{CB} & \textit{Krippendorff's $\alpha$} & \textit{Percent Agreement} & \textit{Spearman's $\rho$} & \textit{Standardized $\alpha$} \\
\midrule
    Bottom-Up-Hebrew    & 0.58 & 0.20 & 0.75 & 0.78 \\
    Bottom-Up-English   & 0.57 & 0.20 & 0.72 & 0.73 \\
    Bottom-Up-Corrected & 0.57 & 0.17 & 0.74 & 0.77 \\
    Top-Down            & 0.49 & 0.13 & 0.70 & 0.72 \\
    Hybrid              & 0.46 & 0.10 & 0.70 & 0.73 \\
    Vanilla DQI         & 0.40 & 0.20 & 0.54 & 0.55 \\
\bottomrule
\end{tabular}

\caption{Inter-Model Agreement (IMA) across six LLMs over 100 transcripts, computed per category on transcript-level scores and averaged within each CB. Percent Agreement counts only exact matches, which are rare between averaged scores; Krippendorff's $\alpha$ weights disagreements by size, Spearman's $\rho$ compares rankings, and Standardized $\alpha$ compares relative scores.}
\label{tab:reliability_scores}
\end{table*}

\subsection{CB Reliability}

\autoref{tab:reliability_scores} presents the reliability evaluation, structured to mirror the evaluation metrics of the original DQI by \citet{steenbergen2003measuring}.
Overall, the IMA results of the CB annotation experiments show some level of agreement for every CB ($\alpha>0$, $\rho>0.5$). The data-driven CBs are consistently the most reliable across workflows, with all three inductive CBs reaching Krippendorff $\alpha$ scores in the range of 0.57--0.58, compared with $\alpha=0.46$ and $\alpha=0.49$ for the \textit{Hybrid} and \textit{Top-Down} CBs, respectively.
In contrast, the baseline Vanilla DQI CB yields the lowest reliability on both rank-based measures ($\alpha=0.40$, $\rho=0.54$, and Standardized $\alpha=0.55$).
hitting the threshold for moderate agreement ($\alpha=0.40$) \citep{Olssonetal2023}. 

Furthermore, when evaluating category-level IMA, the categories associated with the highest values primarily relate to {\em respect} and {\em decorum}, whereas those at the lower end of the reliability spectrum correspond to {\em justification, counterargument engagement}, and {\em emotional regulation}. For example, {\em Respect and Civility} in the Top-Down CB ($\alpha=0.79$) and {\em Procedural Decorum} in the Bottom-Up-English CB ($\alpha=0.83$) exhibit the strongest consensus. Conversely, {\em Justification Level} yields poor agreement in both the Vanilla DQI ($\alpha=0.08$) and Top-Down ($\alpha = 0.20$) CBs, and {\em Emotional Regulation} is the least reliable category of both the Bottom-Up-English ($\alpha=0.15$) and Bottom-Up-Corrected ($\alpha=0.19$) CBs
(see \autoref{tab:reliability_scores_extremes} in \autoref{app_extreame_cat}).

\begin{table}[t]
\centering
\setlength{\tabcolsep}{4pt}
\begin{tabular}{lr}
\toprule
CB & GPT-6-luna ($\rho$) \\
\midrule
Top-Down            & 0.66 \\
Bottom-Up-English   & 0.62 \\
Bottom-Up-Corrected & 0.61 \\
Hybrid              & 0.61 \\
Vanilla DQI         & 0.59 \\
Bottom-Up-Hebrew    & 0.59 \\
\bottomrule
\end{tabular}
\caption{Spearman correlation between each C-DQI CB's discussion-level score and the proxy C-DQI computed by GPT-6-luna, with the default hyperparameters. CB scores are averaged over the six annotator LLMs per transcript; items are all test transcripts ($n=100$). All correlations are significant at $p<10^{-9}$.}
\label{tab:proxy_accuracy}
\end{table}

\subsection{CB Validity}
\label{sec:results_acc}
In \autoref{tab:proxy_accuracy}, we present the Spearman correlations between the discussion-level CB scores and the C-DQI proxy scores across the 100 transcripts. All correlations exceed 0.55 and are statistically significant ($p<10^{-9}$). Notably, the \textit{top-down} CB aligns closest to the proxy ($\rho = 0.66$). This suggests that although the proxy annotators received only minimal binary guidelines, the concept of deliberative quality they apply is itself rooted in the literature, which may favor theory-led CB in the resulting rankings.

\subsection{CB Analysis}

We examine the criteria of each CB to identify their similarities and divergences. 

The only category shared across all CBs is \textit{Respect}, which appears as three separate dimensions in the original DQI alone. According to the DQI, respect divides into respect toward groups in society, respect toward others, and respect toward contradicting arguments. A dominant category in the theory-led CBs is \textit{Constructive Politics}, defined as the ability to bridge disagreements and reach practical solutions, consensus, or compromise. The most similar pair of CBs is the \textit{Hybrid} and \textit{Top-Down}, sharing five out of six metrics.

Notably, specific categories exist exclusively in the deductive workflows, such as participation, level and content of justification, and reciprocity. Conversely, categories appearing solely in the inductive workflows include procedural decorum, emotional regulation, and moderator efficiency.\footnote{For this analysis, the Hybrid workflow is counted as both an inductive and deductive approach.}

\subsection{Discussion}
As shown, the data-driven CBs are more stable across LLM annotation than the theory-based CBs. Specific criteria that appear in the inductive CBs such as emotional regulation and moderator efficiency are likely a product of local cultural nuances in the input data of the CB generation phase.

To observe how participation dynamics manifest, we re-examine the transcripts from \S\ref{sec:data} using the inductive CBs.\footnote{on a 1-4 scale} The {\bf most intensive} meeting is scored 2.5 (Bottom-Up-English) and 2.8 (Bottom-Up-Corrected), while the {\bf most chair-centered} meeting is scored 2.6 and 2.8, respectively. In contrast, the most {\bf evenly distributed} meetings attain the same scores, 3.0 and 3.2 for the respective CBs.
This suggests these CBs favor balanced participation over heavy dominance, inviting further investigation to systematically explore the interplay between structural meeting dynamics and automated discourse quality assessments.

Future directions include applying the C-DQI to analyze fluctuations in deliberative quality across the parliamentary cycle, investigating whether sessions held immediately before recesses exhibit different scores than mid-session meetings and how it systematically varies across the legislative term.

Additionally, the C-DQI enables the comparison of discourse quality analysis across different deliberative environments, such as other parliaments, alternative Israeli legislative bodies, and varying types of Knesset committee protocols.

The results validate the promise of LLM-driven bottom-up strategies for defining and applying discourse quality schemas. Serving as a proof of concept, our study motivates scaling the generation and annotation of CBs. By maneuvering LLM biases to serve as a feature that enhances robustness, rather than viewing them as unwanted artifacts, future research can uncover new analytical schemes that advance concrete legislative goals.

\section{Conclusion}

In this study, we proposed a scalable, LLM-driven methodology for generating and validating discourse quality codebooks. Specifically, we examined four approaches for schema definition, deploying an ensemble of diverse LLMs as zero-shot annotators to evaluate the resulting categories. Our results underscore the value of a bottom-up strategy for constructing reliable discourse evaluation frameworks, with validity comparable to theory-based alternatives. Taken together, this work highlights promising avenues for applying our proposed workflow rather than predefined schemas to diverse practical, political, and social scientific investigations.

\section{Limitations}

We acknowledge the limitations in our study. Although we rely on LLM-inherent biases in our methodology design, we have not examined how these artifacts manifest in annotating the CBs in practice. Additionally, regarding our choice to use segments of 1,500-words: this aims to capture reciprocity across multiple turns, yet larger segments risk reducing interpretability and introducing noise into holistic scoring. This highlights an inherent tension: higher-level abstractions capture broader dynamics but are harder to evaluate against specific criteria. Despite this, holistic evaluation remains valuable for assessing overarching processes and solutions. In future work, we will explore this in more depth. 

Regarding the \textit{Top-Down} CB, the instruction `to develop a discourse quality index' may have biased the model toward the DQI and Habermasian discourse ethics. Further experiments could explore alternative phrasing for the prompt.

Furthermore, the CBs are evaluated separately as a proof of concept; we did not examine their combinations and other balances besides the hybrid. 
Finally, for validity, we rely on statistical correlations. Although a positive correlation indicates the direction of agreement, it doesn't directly translate to accuracy. We acknowledge this as a limitation. Future research may address this by validating the distributional similarity between CB scores and the proxy C-DQI, and can benefit from a manual gold standard as well.

\section{Acknowledgments}

We wish to thank Nadav Oved and Noam Golan from our annotators team for their dedicated research assistance.  This research was supported by the Bar-Ilan University Data Science and Artificial Intelligence Institute, and by the Rector's grant which funded the 2026 interdisciplinary research group on computational analysis of the legislative process. Furthermore, this research was funded by the Israel Science Foundation (grant No. 670/23 and grant No. 2352/26), a KAMIN grant from the Israel Innovation Authority (IIA), and a VATAT grant from the Israeli Planning and Budgeting Committee (PBC), for which we are grateful. We finally wish to thank Google for their generous grant award supporting our lab efforts on AI for Science.

\bibliography{custom}

\appendix

\section{Code-Book Generation Prompts}
\label{appendix_cb_prompts}
\paragraph{Bottom-Up CB. - English}

\begin{mdframed}[backgroundcolor=gray!10, roundcorner=5pt, linewidth=0.5pt]
Attached are 10 raw transcripts from committee meetings in the Israeli parliament, the Knesset.
Using a bottom-up approach, analyze the discourse of these specific transcripts to define metrics that evaluate the quality of discourse in general committee meetings.

Provide:
\begin{enumerate}
    \item A list of the metrics.
    \item The definition of each metric.
\end{enumerate}
\end{mdframed}

\paragraph{Bottom-Up CB. - Hebrew}
\begin{mdframed}[backgroundcolor=gray!10, roundcorner=5pt, linewidth=0.5pt]
\begin{cjhebrew}
lpnyk qb.sy .tqs.t /smkylym prw.twqwlym mww`dwt hknst
\end{cjhebrew}.

\begin{cjhebrew}
hgdr m.tryqwt lh`rkt 'ykwt h/sy.h /sl kl ww`dh    
\end{cjhebrew}.

\begin{cjhebrew}
t.syg 't hm.tryqwt w't hhgdrh /sl kl '.ht mhM    
\end{cjhebrew}
.

\end{mdframed}

\paragraph{Top-Down CB. }
\begin{mdframed}[backgroundcolor=gray!10, roundcorner=5pt, linewidth=0.5pt]
Based on the frameworks in these papers, define metrics to evaluate the quality of discourse in committee transcripts from the Israeli parliament, the Knesset.

Provide:
\begin{enumerate}
    \item
    A list of the metrics.
    \item 
   The definition of each metric. 
\end{enumerate}

\end{mdframed}

\newpage

\paragraph{Hybrid CB. }
\begin{mdframed}[backgroundcolor=gray!10, roundcorner=5pt, linewidth=0.5pt]

Attached are 20 files:
Half are academic literature on democratic deliberation and discourse quality (Theoretical Foundation). The other half are raw committee meeting transcripts from the Israeli parliament, the Knesset (Empirical Context).

Using a combined theory and bottom-up approach, synthesize the academic frameworks and analyze the discourse of these specific transcripts to define metrics that evaluate the quality of discourse in general committee meetings.

Provide:
\begin{enumerate}
    \item 
    A list of the metrics.
    \item 
    The definition of each metric.
\end{enumerate}
\end{mdframed}

\section{Proxy Generation Prompts}
\label{sec:proxy_prompt}
For the proxy generation prompts see \autoref{fig:judge-system-prompt} and \autoref{fig:judge-prompt} for the system and user prompts.

\definecolor{promptbg}{HTML}{F7F8FA}
\definecolor{promptframe}{HTML}{8A9099}
\definecolor{slotcolor}{HTML}{1F5FA8}

\newtcolorbox{promptbox}[2][]{%
  enhanced, breakable,
  colback=promptbg, colframe=promptframe,
  boxrule=0.5pt, arc=2pt,
  left=8pt, right=8pt, top=6pt, bottom=6pt,
  colbacktitle=promptframe!20, coltitle=black,
  fonttitle=\sffamily\bfseries\small,
  title={#2}, #1}

\newcommand{\psec}[1]{%
  \par\medskip\noindent{\sffamily\bfseries\small #1}\par\nobreak\smallskip}
\newcommand{\slot}[1]{\textcolor{slotcolor}{\ttfamily\{#1\}}}

\begin{figure*}[bhp]
\begin{promptbox}{System prompt}
\small
You will review a snippet taken from an Israeli Knesset law bill committee meeting.
The snippet is an exchange of 3 consecutive turns from the transcript, shown with
the turns around it as context. Rate the deliberation quality of the exchange as a
single unit for one deliberative-quality schema.

\psec{Output format}
Return a single valid JSON object and nothing else.
\end{promptbox}
\caption{System prompt used by the LLM judge.}
\label{fig:judge-system-prompt}
\end{figure*}   

\vspace{4pt}

\begin{figure*}[p]
\begin{promptbox}{User prompt (1/2)}
\small
\psec{Context}
Below is a 3-turn exchange taken from a Knesset committee meeting.
Your task is to rate the deliberation quality of the selected exchange.

\psec{Instruction}
Evaluate the overall deliberation quality of the given exchange. The final score
must be binary, where 0 reflects poor or destructive deliberation, and 1 indicates
substantive deliberation.

\begin{itemize}[leftmargin=1.2em, itemsep=2pt, topsep=3pt]
  \item \textbf{Good interactions (code 1):} Statements in which a speaker presents
  an opinion or policy demand backed by logical reasoning, evidence, or
  justification, while maintaining strict decorum, demonstrating a respectful,
  constructive, or appreciative orientation toward other participants and their
  counterarguments.
  \item \textbf{Neutral or poor interactions (code 0):} Statements that express
  unreasoned claims or non-topical small talk (``neutral''), as well as statements
  containing personal attacks, sarcastic degradation, or explicit disrespect toward
  other speakers and their demands (``poor'').
\end{itemize}

\psec{Clarifications}
\begin{itemize}
    [leftmargin=1.6em, itemsep=1pt, topsep=2pt]
  \item A score of 1 requires both elements: (1) respectful discourse, and
  (2) arguments based on internal logic or data.
  \item The definition of 0 includes: procedural statements, cynicism, outbursts,
  or a lack of reasoning, neutral stances and concluding remarks.
  \item Asking or answering a question does not qualify as substantive deliberation
  unless it directly addresses a contested policy issue with reasons or justification.
  \item Do not judge the validity of the claim, only the manner of its presentation;
  we lack the tools to evaluate domain-specific expertise, unless it involves a
  blatant falsehood.
  \item Do not consider the role or identity of the speaker when coding.
  \item Judge the interaction as a whole; even if there is a dominant speaker or
  turn, that role is given weight, but the other speakers are also considered.
  \item Do not infer deliberative quality from surrounding turns. If the exchange
  contains both substantive deliberation and subsequent procedural remarks, score 1
  only when the substantive reasoning is an actual part of the exchange and not
  merely incidental to an otherwise procedural interaction.
  \item Interruptions (marked as ``\texttt{- - -}'') are an indication of a
  disrespectful discussion, but a single interruption is not decisive on its own.
  \item Fragmented exchanges: count a turn as fragmentary if it is cut off
  (``\texttt{- - -}'') or is a short interjection of a few words. If two of the
  three turns are fragmentary, the speakers are not letting each other finish;
  score 0 even if the remaining turn contains a reasoned point.
  \item Implicit cynicism: even if the question is legitimate, the cynical manner
  in which it is presented steers the decision toward 0.
  \item Watch for mocking or sarcastic language, such as belittling a
  professional's request for time by mockingly referring to a basic calculation as
  a ``complex mathematical exercise'', which is considered disrespectful.
  \item Implicit cynicism via rhetorical questioning: If a speaker undermines a
  proposal, explanation, or request for authority by using a rhetorical,
  challenging question (e.g., mockingly asking ``Until now you didn't need
  this?''), it is considered cynical and disrespectful. Score the exchange as 0,
  even if the challenged speaker responds in a substantive and professional manner.
  \item Use the provided surrounding context to identify cynicism in the exchange
  to rate.
  \item If a participant asks for a date, deadline, or concrete commitment, or
  objects that a timeline is too long, a general explanation of the process or a
  vague promise (e.g.\ ``soon'') without providing that information should score 0
  for non-responsiveness.
  \item A reasoned procedural objection about notice, timing, or fairness does not
  by itself qualify as good deliberation. To receive a score of 1, the exchange must
  also include substantive engagement with the objection, such as a direct answer,
  justification, or concrete accommodation.

\end{itemize}

\end{promptbox}
\caption{User prompt used by the LLM judge (continued on the next page).}
\label{fig:judge-prompt}
\end{figure*}

\begin{figure*}[t!]
\ContinuedFloat
\begin{promptbox}{User prompt (2/2)}
\small

\begin{itemize}
    \item Merely reporting an administrative position, explaining implementation,
  announcing a vote, expressing agreement, or closing the discussion does not
  qualify as substantive deliberation unless it directly addresses a contested
  policy issue with reasons or justification. A respectful but purely procedural or
  informational exchange therefore receives 0.
  \item Distinguish \emph{why} from \emph{how}. Explaining why a bill, provision,
  or request is needed (its purpose, the problem it solves, the consequences of not
  acting) is justification and can support a 1. Describing how a provision will be
  implemented, what an agency will check, or what should be calculated or requested
  from another body (action items) is informational and scores 0.
  \item An explanation that resolves a concern or confusion raised by another
  participant, and is acknowledged by them, is responsive engagement and can
  support a 1.
   
\end{itemize}

\psec{Output format}
Return only a valid JSON object with this exact structure:
\begin{Verbatim}[fontsize=\footnotesize, xleftmargin=1.2em]
{
  "rationale": "<1-2 sentences justifying the score>",
  "score": <1 or 0>
}
\end{Verbatim}

\psec{Metadata}
Knesset \slot{number} \,|\, \slot{committee} \,|\, \slot{date} \,|\, \slot{topics discussed}

\psec{Context before the exchange}
\slot{context before}

\psec{Exchange to rate}
\slot{exchange}

\psec{Context after the exchange}
\slot{context after}
\end{promptbox}
\caption{User prompt used by the LLM judge (continued). Text in
\textcolor{slotcolor}{\ttfamily\{blue\}} is filled in per exchange.}
\end{figure*}

\begin{table*}[h]
\centering
  \begin{tabular}{llrlr}
    \toprule
    & \multicolumn{2}{c}{Most reliable feature}
    & \multicolumn{2}{c}{Least reliable feature} \\
    \cmidrule(lr){2-3} \cmidrule(lr){4-5}
    Codebook & Feature & $\alpha$ & Feature & $\alpha$ \\
    \midrule
    Vanilla DQI         & Respect toward Groups     & 0.60
                        & Justification Level       & 0.08 \\
    Top-Down          & Respect and Civility      & 0.79
                        & Justification Level       & 0.20 \\
    Bottom-Up-English   & Procedural Decorum        & 0.83
                        & Emotional Regulation      & 0.15 \\
    Bottom-Up-Corrected & Procedural Decorum        & 0.80
                        & Emotional Regulation      & 0.19 \\
    Bottom-Up-Hebrew    & Decorum and Tone          & 0.80
                        & Chair Management Efficiency & 0.36 \\
    Hybrid              & Civility and Conflict Mgmt.\ & 0.69
                        & Moderator Efficacy        & 0.22 \\
    \bottomrule
  \end{tabular}
\caption{Most and least reliable features per C-DQI codebook, determined by highest and lowest Krippendorff's $\alpha$ scores.
}
\label{tab:reliability_scores_extremes}
\end{table*}

\section{Extreme Category IMA}
\label{app_extreame_cat}

\autoref{tab:reliability_scores_extremes} describes the categories in each CB that exhibit the highest and lowest IMA scores.

\section{Constructing the Bottom-Up-Corrected Codebook}
\label{cb_corrections}

We constructed the corrected inductive codebook by merging the English and Hebrew codebooks, retaining every distinct metric from each. Where the two versions overlapped, we consolidated them into a single metric; Hebrew’s Decorum and Tone category was split between Procedural Decorum and Respectful Engagement. We removed most illustrative examples (``such as…'') to avoid biasing human or model annotators toward specific cases. We also adopted the bullet structure of the Hebrew codebook and standardized it so that every metric has two labeled poles, ``High'' and ``Low''. Where a source codebook described only one pole, we defined the missing one to be the parallel opposite of the existing one. Finally, for the Procedural Decorum metric, we replaced ``unanimous agreement'' from the English codebook with ``respectful disagreement'', since agreement is not a condition for good decorum.

\section{CodeBooks}
\label{sec:appen_codebooks}
\paragraph{Vanilla DQI}

\begin{itemize}[leftmargin=*]
    \item[] \textbf{0. Demand} \\
    A proposal about what should or should not be decided or done. A demand may support changing, adopting, rejecting, or preserving a policy or course of action.
    
    \item[] \textbf{1. Participation} \\
    \textbf{Definition:} Evaluates whether a participant can complete a relevant contribution without being prevented from speaking by disruption or formal intervention. \\
    \textbf{Coding Categories / Scale:}
    \begin{itemize}[leftmargin=*]
        \item[] \textbf{Code 0 (Participation Interrupted):} The speaker explicitly indicates that an interruption is disturbing or preventing the contribution, or the contribution is terminated by a formal decision. Do not assign this code merely because the allocated speaking time has expired.
        \item[] \textbf{Code 1 (Normal Participation Possible):} The participant is able to make the relevant contribution under ordinary speaking conditions.
    \end{itemize}

    \item[] \textbf{2. Level of Justification for Demands} \\
    \textbf{Definition:} Measures the completeness of the reasons supplied for a demand. The substantive truth or persuasiveness of the argument is not judged. A complete justification contains (1) a conclusion about what should or should not be done, (2) a supporting reason, and (3) an explicit or unmistakably implied inferential link between the reason and conclusion. \\
    \textbf{Coding Categories / Scale:}
    \begin{itemize}[leftmargin=*]
        \item[] \textbf{Code 0 (No Justification):} A demand or evaluation is stated, but no supporting reason is provided.
        \item[] \textbf{Code 1 (Inferior Justification):} A supporting consideration is mentioned, but the link between the reason and the demand is incomplete or unclear. A conclusion supported only by illustration also receives Code 1.
        \item[] \textbf{Code 2 (Qualified Justification):} At least one complete justification connects a demand to a reason through an explicit or clearly implied inferential link.
        \item[] \textbf{Code 3 (Sophisticated Justification):} At least two distinct complete justifications are supplied, whether for the same demand or for different demands within the speech act.
    \end{itemize}
    \textbf{Decision Rule:} Assign the highest level whose requirements are fully met. An implicit linkage counts only when its intended meaning would be beyond reasonable doubt to participants in the debate.

    \item[] \textbf{3. Content of Justification for Demands} \\
    \textbf{Definition:} Identifies whose interests are explicitly invoked when a speaker justifies a demand: a constituency or particular group, society collectively, or disadvantaged groups under a fairness-based conception of the common good. \\
    \textbf{Coding Categories:}
    \begin{itemize}[leftmargin=*]
        \item[] \textbf{Code 0 (Constituency or Group Interest):} At least one explicit appeal is made to the interest of a particular constituency, organization, sector, party, region, or social group.
        \item[] \textbf{Code 1 (Neutral Statement):} There is no explicit reference to a constituency or group interest and no explicit appeal to either form of the common good.
        \item[] \textbf{Code 2 (Common-Good Orientation):} The speaker explicitly justifies a demand using at least one of the following forms of common-good reasoning:
        \begin{itemize}[leftmargin=*]
            \item[] \textbf{Utilitarian or Collective Common Good:} An appeal to the best outcome for the greatest number, society as a whole, or a larger collective such as Europe or humanity.
            \item[] \textbf{Difference-Principle Common Good:} An appeal to protecting or helping the least advantaged, poorest, most discriminated against, or otherwise vulnerable members of society, including children and future generations, or an appeal to fairness and justice in this sense.
        \end{itemize}
    \end{itemize}
    \textbf{Decision Rule:} Assign the highest level whose requirements are fully met.

    \item[] \textbf{4. Respect toward Groups to Be Helped (Empathy)} \\
    \textbf{Definition:} Evaluates how the speaker characterizes the people or groups who would be affected or helped by the demand. \\
    \textbf{Coding Categories / Scale:}
    \begin{itemize}[leftmargin=*]
        \item[] \textbf{Code 0 (No Respect):} At least one negative or degrading statement is made about a relevant group, and no positive statement is made. A single negative statement is sufficient.
        \item[] \textbf{Code 1 (Neutral):} Neither negative nor explicitly positive statements are made about the relevant groups.
        \item[] \textbf{Code 2 (Explicit Respect):} At least one explicitly positive, valuing, empathetic, or dignity-affirming statement is made about a relevant group. This code still applies when negative statements also occur.
    \end{itemize}
    \textbf{Priority Rule:} Explicit respect takes priority over negativity: mixed positive and negative statements receive Code 2.

    \item[] \textbf{5. Respect toward the Demands of Others} \\
    \textbf{Definition:} Evaluates whether a speaker degrades, treats neutrally, or explicitly values demands advanced by other participants. \\
    \textbf{Coding Categories / Scale:}
    \begin{itemize}[leftmargin=*]
        \item[] \textbf{Code 0 (No Respect):} At least one negative or degrading statement is made toward another demand, or toward the person or group advancing it, and no positive statement is made.
        \item[] \textbf{Code 1 (Neutral):} Other demands receive neither negative nor explicitly positive treatment.
        \item[] \textbf{Code 2 (Explicit Respect):} At least one positive statement values any aspect of another participant's demand. This code still applies when negative statements also occur.
        \item[] \textbf{N/A (Not Applicable):} Only one demand is on the agenda and the speaker supports it; consequently, there is no distinct competing demand to evaluate.
    \end{itemize}
    \textbf{Priority Rule:} Explicit respect takes priority over negativity: mixed positive and negative statements receive Code 2. Criticism of another demand may additionally be coded under Level and Content of Justification when it functions as part of the speaker's own argument.

    \item[] \textbf{6. Respect toward Counterarguments} \\
    \textbf{Definition:} Evaluates how a speaker responds to arguments that contradict the speaker's conclusion concerning a demand. The category concerns engagement with opposing reasons, not merely interpersonal politeness. \\
    \textbf{Coding Categories / Scale:}
    \begin{itemize}[leftmargin=*]
        \item[] \textbf{Code 0 (Counterarguments Ignored):} One or more counterarguments are already available, but the speaker does not address any of them.
        \item[] \textbf{Code 1 (Included but Explicitly Degraded):} Counterarguments are acknowledged, but at least one is degraded, mocked, or attacked, and none is explicitly valued. Attacks on the people or groups advancing the counterarguments also count.
        \item[] \textbf{Code 2 (Included Neutrally):} Counterarguments are acknowledged or addressed without either negative degradation or explicit positive valuation.
        \item[] \textbf{Code 3 (Included and Explicitly Valued):} At least one aspect of at least one counterargument is explicitly recognized as valid, correct, useful, or worthy of consideration. This code still applies when other negative statements occur.
        \item[] \textbf{N/A (Not Applicable):} No counterargument has yet been introduced and the speaker does not anticipate one.
    \end{itemize}
    \textbf{Totality Rule:} When several counterarguments exist, assign one code to the speaker's treatment of the set as a whole. Explicit valuation takes priority over negative treatment; without valuation, any degradation leads to Code 1. An anticipated counterargument should receive an actual code rather than N/A.

    \item[] \textbf{7. Constructive Politics} \\
    \textbf{Definition:} Measures whether the speaker remains fixed in a position or advances a mediating proposal, distinguishing proposals that address the present agenda from alternatives that belong to another policy agenda. \\
    \textbf{Coding Categories / Scale:}
    \begin{itemize}[leftmargin=*]
        \item[] \textbf{Code 0 (Positional Politics):} The speaker maintains a position without offering a mediating or compromise proposal.
        \item[] \textbf{Code 1 (Alternative Proposal):} The speaker advances a proposal that could mediate the broader conflict but does not fit the current agenda or resolve the decision presently under debate. It would need to be considered under another agenda.
        \item[] \textbf{Code 2 (Mediating Proposal):} The speaker offers a concrete compromise or accommodation within the current agenda that attempts to bridge competing concerns or modify the proposal under consideration.
    \end{itemize}
    \textbf{Coding Note:} A proposal is not mediating merely because it differs from the current proposal. It must attempt to accommodate competing concerns; the distinction between Codes 1 and 2 depends on whether it can be acted upon within the present agenda.
\end{itemize}

\paragraph{Bottom-Up CB. - English}
\begin{itemize}[leftmargin=*]
    \item[] \textbf{Metric 1: Procedural Decorum} \\
    This metric evaluates the extent to which participants follow parliamentary rules, wait for their turn to speak, and avoid shouting or talking over one another. In some committees, the chairperson must repeatedly threaten to remove members to maintain order. Conversely, other meetings proceed with unanimous agreement, polite praises, and uninterrupted presentations.

    \item[] \textbf{Metric 2: Topic Adherence} \\
    This metric measures the ratio of substantive discussion regarding the agenda versus unrelated political grandstanding or personal attacks. High topic adherence is observed when members strictly debate specific legal definitions, such as defining security offenses versus treason. Low adherence occurs when speaking time meant for legislative reservations is entirely used to attack political opponents on unrelated national issues.

    \item[] \textbf{Metric 3: Constructive Problem-Solving} \\
    This metric assesses the ability of committee members and government officials to bridge disagreements and reach practical legislative compromises. This metric is positive when members negotiate temporary orders to test a law's impact before making it permanent. It is also evident when legal advisors and ministries successfully negotiate timelines for administrative tasks, such as extending a deadline from three months to seven months to accommodate workload constraints.

    \item[] \textbf{Metric 4: Respectful Engagement} \\
    This metric analyzes the tone and language used when interacting with civil servants, academic experts, and members of the public. Discourse quality drops when elected officials use dismissive language, such as calling opposing arguments ``bullshit''. It also drops when there is friction between politicians and civil servants, such as when an official allegedly calls a member a ``populist''. High quality is shown when chairpersons actively listen to and validate the emotional testimonies of local mayors.

    \item[] \textbf{Metric 5: Emotional Regulation} \\
    This metric examines how the committee handles highly charged emotional topics or national tragedies while conducting legislative business. Poor emotional regulation is visible when procedural duties, like budget discussions, clash with days of national mourning, leading to accusations of callousness and ``disgusting'' behavior between members. High emotional tension is also present when bereaved parents confront lawmakers directly during security-related debates.
\end{itemize}


{\par\noindent\beginL\textnormal{\textbf{Bottom-Up CB. - Hebrew}}\endL\par}

\vspace{10pt}
\begin{hpar}{\Large \EN{\textbf{1.}}\; `nyynywt wmyqwd bdywN \EN{\textbf{(Relevance and Focus)}}}\end{hpar}
\begin{hpar}{\large hgdrh\EN{:}}\; mdd hbw.hN `d kmh hdwbryM \EN{(}.hbry hknst whmwzmnyM\EN{) }mtmqdyM bnw/s' hmq.sw`y l/smw htknsh hww`dh\EN{, }l`wmt gly/sh lpylybs.tryM\EN{, }h`l't nw/s'yM /s'ynM q/swryM\EN{, }'w n'wmyM pwly.tyyM\EN{.}\end{hpar}
\vspace{3pt}
\begin{hpar}\EN{\textbullet}\; {\large 'ykwt /sy.h gbwhh\EN{:}}\; htmqdwt bs`ypy h.hwq\EN{, }nysw.hyM m/sp.tyyM wptrwN .hsmyM byrwqr.tyyM\EN{.}\end{hpar}
\vspace{3pt}
\begin{hpar}\EN{\textbullet}\; {\large 'ykwt /sy.h nmwkh\EN{:}}\; /symw/s bzmN hww`dh ln'wmyM pwly.tyyM 'rwkyM /s'ynM q/swryM ls`yP hndwN kll\EN{, }kpy /snytN lr'wt bhg/st `/srwt hstyygwywt wnymwqN `l ydy tqypt h/sr lby.t.hwN l'wmy bnw/s'yM /swnyM l.hlw.tyN \EN{(}kgwN t'wnwt drkyM 'w gz`nwt\EN{) }bmhlK dywN `l .hwq l/skt `wrky hdyN\EN{.}\end{hpar}
\vspace{10pt}
\begin{hpar}{\Large \EN{\textbf{2.}}\; trbwt hdywN wsgnwN hhtb.t'wt \EN{\textbf{(Decorum and Tone)}}}\end{hpar}
\begin{hpar}{\large hgdrh\EN{:}}\; mdd hm`ryK 't rmt hkbwd hhddy\EN{, }hhq/sbh w'wpN hhtb.t'wt byN hm/sttpyM\EN{. }hmdd nb.hN `l ydy spyrt qry'wt bynyyM\EN{, }htpr.swywt ldbry '.hryM\EN{, }hrmt qwl\EN{, }w/symw/s b/sph bw.th 'w tqypwt 'y/sywt\EN{.}\end{hpar}
\vspace{3pt}
\begin{hpar}\EN{\textbullet}\; {\large 'ykwt /sy.h gbwhh\EN{:}}\; hq/sbh ml'h ldwbryM\EN{, }hb`t h`rkh hddyt gM byN qw'ly.syh l'wpwzy.syh\EN{, }wdywN /slyw wmq.sw`y\EN{.}\end{hpar}
\vspace{3pt}
\begin{hpar}\EN{\textbullet}\; {\large 'ykwt /sy.h nmwkh\EN{:}}\; htpr.swywt tkwpwt\EN{, }qw/sy lh/slyM m/sp.t\EN{, }w/symw/s bby.twyyM pwg`nyyM wh'/smwt 'y/sywt \EN{(}kgwN .hylwpy hh'/smwt \EN{"}twmky .trwr\EN{" }w\EN{-"}gys .hmy/sy\EN{" }'l mwl \EN{"}mp`yl .trwr\EN{" }w\EN{-"}p'/sys.tyM\EN{").}\end{hpar}
\vspace{10pt}
\begin{hpar}{\Large \EN{\textbf{3.}}\; y`ylwt nyhwl hy/sybh \EN{\textbf{(Efficiency of Chair Management)}}}\end{hpar}
\begin{hpar}{\large hgdrh\EN{:}}\; mdd hbw.hN 't ykwltw /sl yw/sb\EN{-}r'/s hww`dh l/slw.t bdywN\EN{, }lntb 'wtw lhkr`wt m`/sywt\EN{, }lq.tw` hpr`wt wlmnw` mhdywN l.s't mkll /sly.th\EN{, }twK /smyrh `l hwgnwt klpy hm/sttpyM\EN{.}\end{hpar}
\vspace{3pt}
\begin{hpar}\EN{\textbullet}\; {\large nyhwl h.hl.ty w'q.tyby\EN{:}}\; ykwlt /sl hyw/sb\EN{-}r'/s lq.tw` wykw.hyM byrwqr.tyyM 'rwkyM\EN{, }lhn.hwt 't gwrmy hmq.sw` lhgy` lp/srwt prq.tywt\EN{, }wlqbl h.hl.twt `l hmqwM twK nzyph bgwrmyM /sl' mb.s`yM 't `bwdtM kr'wy \EN{(}kmw hdry/sh lqnws r/swywt mqwmywt /s/swl.hwt qnswt /sww' l'mbwlnsyM\EN{).}\end{hpar}
\vspace{3pt}
\begin{hpar}\EN{\textbullet}\; {\large qw/sy bnyhwl hdywN\EN{:}}\; 'wbdN /sly.th `l sdr hdwbryM\EN{, }nysywnwt .hwzryM wn/snyM \EN{(}wl`ytyM kw/slyM\EN{) }lh/sqy.t qry'wt bynyyM\EN{, }w'ywmyM tkwpyM bhw.s't m/sttpyM ll' h.sl.hh\EN{.}\end{hpar}
\vspace{10pt}
\begin{hpar}{\Large \EN{\textbf{4.}}\; mq.sw`ywt whstmkwt `l ntwnyM \EN{\textbf{(Professionalism and Data Reliance)}}}\end{hpar}
\begin{hpar}{\large hgdrh\EN{:}}\; mdd hbwdq 't mydt hhy/s`nwt /sl m/sttpy hww`dh `l `wbdwt\EN{, }ntwnyM\EN{, }m.hqryM \EN{(}kgwN mrkz hmyd` whm.hqr /sl hknst\EN{), }whq/sbh lyy`w.S hm/sp.ty wldrgyM hmq.sw`yyM bm/srdy hmm/slh\EN{.}\end{hpar}
\vspace{3pt}
\begin{hpar}\EN{\textbullet}\; {\large 'ykwt /sy.h gbwhh\EN{:}}\; nytw.h mdwqdq /sl h.hwq s`yP '.hr s`yP y.hd `M hyy`w.S hm/sp.ty\EN{, }twK hbnt hh/slkwt hm/sp.tywt /sl kl mylh b.hwq \EN{(}kgwN hhbdl byN \EN{"}.hwq y.hwl\EN{" }l\EN{-"}hgblwt l' y.hwlw\EN{").}\end{hpar}
\vspace{3pt}
\begin{hpar}\EN{\textbullet}\; {\large p/srh mq.sw`yt mbwsst ntwnyM\EN{:}}\; qby`t lw.hwt zmnyM ry'lyyM \EN{(}kgwN wykw.h `l mspr h.hwd/syM hndr/s lww`dt /s.hrwryM l`bwr `l \EN{120 }tyqyM plylyyM\EN{) }mwl gwrmy hmq.sw` bhnhlt bty hm/sp.t wm/srd hm/sp.tyM\EN{.}\end{hpar}
\vspace{10pt}
\begin{hpar}{\Large \EN{\textbf{5.}}\; 'qlyM rg/sy wgylwy 'mptyh \EN{\textbf{(Emotional Climate and Empathy)}}}\end{hpar}
\begin{hpar}{\large hgdrh\EN{:}}\; mdd hbw.hN 't ykwlt hww`dh lhkyl wlhgyb lm.sbyM rg/syyM mwrkbyM\EN{, }nwk.hwtN /sl m/sp.hwt /skwlwt wm/sp.hwt .h.twpyM\EN{, }whht'mh /sl qywM hdywN lm.sy'wt h.hy.swnyt \EN{(}'yrw`yM by.t.hwnyyM q/syM\EN{).}\end{hpar}
\vspace{3pt}
\begin{hpar}\EN{\textbullet}\; {\large .hwsr hlymh rg/syt\EN{/}mty.hwt q/sh\EN{:}}\; qywM dywN \EN{"}rgyl\EN{" }/sm`wrr z`M bqrb m/sttpyM `l rq` 'yrw`yM by.t.hwnyyM q/syM bzmN 'mt\EN{, }mh /smwbyl lpy.sw.S hdywN wlh.t.ht h'/smwt `l '.tymwt wnytwq \EN{("}`rlwt lb\EN{").}\end{hpar}
\vspace{3pt}
\begin{hpar}\EN{\textbullet}\; {\large mtN mqwM lk'b\EN{:}}\; `.syrt /s.tP h.hqyqh kdy l'p/sr lm/sp.hwt .h.twpyM wm/sp.hwt /skwlwt lh/smy` 't qwlN w't z`qtN\EN{, }whb`t hzdhwt w'mptyh m.sd .hbry hknst whyw/sb\EN{-}r'/s lm.swqtM\EN{.}\end{hpar}

\paragraph{Bottom-up CB - Corrected}

\begin{itemize}[leftmargin=*]
    \item[] \textbf{Metric 1: Procedural Decorum} \\
    Evaluates the extent to which participants follow parliamentary rules, wait for their turn to speak, and avoid shouting or talking over one another.
    \begin{itemize}[leftmargin=*]
        \item[] \textbf{High Decorum:} Observed in meetings that proceed with polite exchanges, uninterrupted presentations, and respectful disagreement.
        \item[] \textbf{Low Decorum:} Associated with committees where the chairperson must repeatedly threaten to remove members to maintain order.
    \end{itemize}

    \item[] \textbf{Metric 2: Topic Adherence} \\
    Measures the ratio of substantive discussion regarding the agenda versus unrelated political grandstanding or personal attacks.
    \begin{itemize}[leftmargin=*]
        \item[] \textbf{High Adherence:} Observed when the majority of participants strictly stick to the discussion agenda, debating specific outcomes and legal definitions.
        \item[] \textbf{Low Adherence:} Observed when discussions frequently go off-topic, or when speaking time meant for legislative reservations is entirely used to attack political opponents on unrelated national issues.
    \end{itemize}

    \item[] \textbf{Metric 3: Constructive Problem-Solving} \\
    Assesses the ability of committee members and government officials to bridge disagreements and reach practical legislative compromises.
    \begin{itemize}[leftmargin=*]
        \item[] \textbf{High Constructiveness:} Indicated by members negotiating temporary orders to test a law's impact before making it permanent, or when legal advisors and ministries successfully negotiate timelines for administrative tasks.
        \item[] \textbf{Low Constructiveness:} Indicated by members adopting zero-sum ideological positions that lead to legislative deadlocks, or government ministries engaging in jurisdictional turf wars and blame-shifting instead of finding actionable paths forward.
    \end{itemize}

    \item[] \textbf{Metric 4: Respectful Engagement} \\
    Analyzes the tone and language used when interacting with civil servants, academic experts, and members of the public.
    \begin{itemize}[leftmargin=*]
        \item[] \textbf{High Respect:} Demonstrated when participants utilize active listening, validate each other's opinions, and maintain a professional tone.
        \item[] \textbf{Low Respect:} Characterized by friction between participants and the use of dismissive or vulgar language (e.g., calling opposing arguments ``bullshit'').
    \end{itemize}

    \item[] \textbf{Metric 5: Emotional Regulation} \\
    Examines how the committee handles highly charged emotional topics or national tragedies while conducting legislative business.
    \begin{itemize}[leftmargin=*]
        \item[] \textbf{High Regulation:} Demonstrated when lawmakers maintain composure, empathy, and respect during emotionally charged testimonies. The committee successfully balances sensitivity to national tragedies with its legislative duties without allowing the discussion to devolve into chaos.
        \item[] \textbf{Low Regulation:} Demonstrated when routine business clashes with sensitive events, resulting in insults, accusations of callousness, or uncontrolled tension when lawmakers are confronted by grieving citizens.
    \end{itemize}

    \item[] \textbf{Metric 6: Meeting Management Effectiveness} \\
    Examines the committee chairperson's ability to control the discussion, steer it toward practical decisions, cut off disruptions, and prevent the debate from spiraling out of control, all while maintaining fairness.
    \begin{itemize}[leftmargin=*]
        \item[] \textbf{High Effectiveness:} The chairperson successfully cuts short lengthy bureaucratic arguments, guides professional entities to reach practical compromises, makes on-the-spot decisions, and holds parties accountable for failing to perform their duties.
        \item[] \textbf{Low Effectiveness:} Characterized by a loss of control over the speaking order, repeated attempts to quiet heckling, and frequent but unsuccessful threats to remove participants from the room.
    \end{itemize}

    \item[] \textbf{Metric 7: Professionalism and Reliance on Data} \\
    Assesses the extent to which committee participants rely on facts, data, and research, and their attentiveness to professional echelons and legal counsel.
    \begin{itemize}[leftmargin=*]
        \item[] \textbf{High Professionalism:} Characterized by a meticulous, section-by-section analysis of the law alongside legal counsel, demonstrating an understanding of the legal implications of the legislation. Compromises set realistic timetables in coordination with professional authorities.
        \item[] \textbf{Low Professionalism:} Characterized by participants ignoring research and data, disregarding professional or legal advice, and arguing over legislation without understanding its technical or practical implications.
    \end{itemize}
\end{itemize}

\paragraph{Top-Down CB.}

\begin{itemize}[leftmargin=*]
    \item[] \textbf{1. Level of Justification (Argumentative Rationality)} \\
    \textbf{Theoretical Origin:} Discourse Quality Index (Steenbergen et al., 2003); Stromer-Galley (2007); Friess \& Eilders (2015). \\
    \textbf{Definition:} Evaluates the logical structure and completeness of reasons provided by Members of the Knesset (MKs) or invited witnesses when advancing a policy proposal, amendment, or claim. \\
    \textbf{Coding Categories / Scale:}
    \begin{itemize}[leftmargin=*]
        \item[] \textbf{Level 0 (No Justification):} A demand/claim is stated without any supporting reason.
        \item[] \textbf{Level 1 (Inferior Justification):} A reason is stated, but the logical link between the demand and the reason is incomplete or relies strictly on illustration.
        \item[] \textbf{Level 2 (Qualified Justification):} At least one complete, explicit, or clearly implied inferential link connects the demand to a supporting reason.
        \item[] \textbf{Level 3 (Sophisticated Justification):} At least two distinct, complete inferential links are provided (either multiple complete reasons for one demand or reasons spanning multiple demands).
    \end{itemize}

    \item[] \textbf{2. Content of Justification (Common Good Orientation)} \\
    \textbf{Theoretical Origin:} Steenbergen et al. (2003); Habermas (1996); Cohen (1989). \\
    \textbf{Definition:} Assesses whether a speaker justifies claims solely through narrow sectoral/party/constituency interests or frames arguments in terms of broader societal benefit. \\
    \textbf{Coding Categories / Scale:}
    \begin{itemize}[leftmargin=*]
        \item[] \textbf{Group/Constituency Interest:} Explicit appeals to narrow political, regional, or constituency benefits.
        \item[] \textbf{Neutral Statement:} No explicit reference to either narrow group interests or the broader common good.
        \item[] \textbf{Common Good (Utilitarian):} Explicit appeal to the best outcome for Israeli society as a whole or the greatest benefit for the majority.
        \item[] \textbf{Common Good (Difference Principle):} Explicit appeal to protecting or helping the least advantaged, vulnerable, or marginalized groups in society (Rawlsian difference principle).
    \end{itemize}

    \item[] \textbf{3. Sourcing and Evidence Base} \\
    \textbf{Theoretical Origin:} Stromer-Galley (2007); Friess \& Eilders (2015). \\
    \textbf{Definition:} Measures the extent to which participants support their positions using verifiable external references, expert testimonies, or official materials rather than unsupported assertions. \\
    \textbf{Coding Categories:}
    \begin{itemize}[leftmargin=*]
        \item[] \textbf{Official/Committee Documents:} Reference to Knesset Research and Information Center (MMM) reports, ministry briefings, or draft bill texts.
        \item[] \textbf{External/Expert Sources:} Reference to academic research, statistical data, court rulings, or international benchmarks.
        \item[] \textbf{Mass Media:} Reference to news reports, journalism, or public media broadcasts.
        \item[] \textbf{Personal Experience / Narrative:} Reference to first-hand accounts, constituent stories, or personal anecdotes.
        \item[] \textbf{No Source:} Claim made without grounding in external or experiential sources.
    \end{itemize}

    \item[] \textbf{4. Interactivity and Reciprocity} \\
    \textbf{Theoretical Origin:} Stromer-Galley (2007); Friess \& Eilders (2015). \\
    \textbf{Definition:} Evaluates whether committee dialogue constitutes a genuine, reciprocal exchange of ideas where speakers listen and respond to one another, or consists of disconnected monologues (``soap-boxing''). \\
    \textbf{Coding Indicators:}
    \begin{itemize}[leftmargin=*]
        \item[] \textbf{Formal Interactivity:} Whether a speech turn explicitly addresses a prior speaker or turn.
        \item[] \textbf{Substantive Interactivity:} Whether a speaker directly addresses, rebuts, clarifies, or builds upon the specific arguments made by a prior participant.
        \item[] \textbf{Information-Seeking / Questioning:} Presence of genuine questions aimed at eliciting clarification or facts from witnesses, government officials, or fellow MKs (distinguished from rhetorical questions).
        \item[] \textbf{Clarification (Self/Other):} Speech acts that explicitly clarify a speaker’s own prior position or summarize another participant’s argument to verify understanding.
    \end{itemize}

    \item[] \textbf{5. Respect and Civility toward Actors and Demands} \\
    \textbf{Theoretical Origin:} Steenbergen et al. (2003); Young (1996); Friess \& Eilders (2015). \\
    \textbf{Definition:} Measures the interpersonal tone and level of mutual recognition accorded to other participants, social groups, and policy proposals during committee proceedings. \\
    \textbf{Coding Categories / Scale:}
    \begin{itemize}[leftmargin=*]
        \item[] \textbf{Respect toward Groups:} Evaluates whether social, ethnic, or political groups affected by the bill are degraded (0), treated neutrally (1), or explicitly valued/respected (2).
        \item[] \textbf{Respect toward Demands:} Evaluates whether policy proposals/amendments raised by fellow participants are degraded/insulted (0), treated neutrally (1), explicitly valued (2), or agreed with (3).
        \item[] \textbf{Civility / Disruption:} Tracks presence of personal insults, ad hominem attacks, or unprocedural interruptions.
    \end{itemize}

    \item[] \textbf{6. Engagement with Counterarguments} \\
    \textbf{Theoretical Origin:} Steenbergen et al. (2003); Bächtiger et al. (2010). \\
    \textbf{Definition:} Assesses how speakers handle opposing viewpoints or counterarguments raised by coalition/opposition members or invited experts. \\
    \textbf{Coding Categories / Scale:}
    \begin{itemize}[leftmargin=*]
        \item[] \textbf{Level 0 (Ignored):} Counterarguments exist on the table, but the speaker completely ignores them.
        \item[] \textbf{Level 1 (Included but Degraded):} Counterarguments are acknowledged but explicitly degraded, mocked, or dismissed without substantive engagement.
        \item[] \textbf{Level 2 (Included Neutrally):} Counterarguments are acknowledged and restated neutrally without personal or thematic degradation.
        \item[] \textbf{Level 3 (Included and Valued):} Counterarguments are acknowledged and explicitly praised or treated as valid considerations, even if ultimately rejected or integrated.
    \end{itemize}

    \item[] \textbf{7. Constructive Politics and Compromise Orientation} \\
    \textbf{Theoretical Origin:} Steenbergen et al. (2003); Bächtiger et al. (2010); Friess \& Eilders (2015). \\
    \textbf{Definition:} Measures whether committee participants remain locked in rigid, zero-sum partisan stances or actively attempt to build consensus and formulate acceptable compromises. \\
    \textbf{Coding Categories / Scale:}
    \begin{itemize}[leftmargin=*]
        \item[] \textbf{Positional Politics (Level 0):} Speakers rigidly maintain initial party lines without attempt at compromise, accommodation, or reconciliation.
        \item[] \textbf{Alternative Proposal (Level 1):} A mediating proposal is offered that belongs to a different agenda or jurisdiction, failing to resolve the immediate conflict on the current draft text.
        \item[] \textbf{Mediating Proposal (Level 2):} A speaker advances a concrete, integrative compromise or amendment that directly fits the current agenda and seeks to bridge coalition-opposition or stakeholder divides.
    \end{itemize}

    \item[] \textbf{8. Topic Coherence and Agenda Focus} \\
    \textbf{Theoretical Origin:} Stromer-Galley (2007); Friess \& Eilders (2015). \\
    \textbf{Definition:} Evaluates whether committee discussion stays focused on the legislative agenda (structuring topic) or drifts into uncoordinated, off-topic partisan disputes (interactional topic drift). \\
    \textbf{Coding Indicators:}
    \begin{itemize}[leftmargin=*]
        \item[] \textbf{On-Topic (Structuring):} Statements directly addressing the bill, policy options, or specific committee agenda items.
        \item[] \textbf{On-Topic (Related Problem):} Statements addressing the broader policy issue (e.g., educational quality, defense budget) without directly addressing the specific legislative choices at hand.
        \item[] \textbf{Off-Topic / Drift:} Statements digressing into unrelated political scandals, electoral grandstanding, or personal disputes unlinked to the agenda.
    \end{itemize}

    \item[] \textbf{9. Participatory Equality and Vocal Distribution} \\
    \textbf{Theoretical Origin:} Stromer-Galley (2007); Knight \& Johnson (1997); Friess \& Eilders (2015). \\
    \textbf{Definition:} Measures whether speaking opportunities and floor time are equitably distributed among participants or monopolized by committee chairs, coalition members, or dominant speakers. \\
    \textbf{Coding Indicators:}
    \begin{itemize}[leftmargin=*]
        \item[] \textbf{Turn \& Word-Count Distribution:} Quantitative balance of speech turns and word volume across coalition vs. opposition MKs, and committee members vs. invited external experts/citizens.
        \item[] \textbf{Procedural Equal Access:} Absence of arbitrary silencing, unequal time enforcement by the chair, or procedural coercion that prevents backbenchers or opposition members from presenting amendments.
    \end{itemize}

    \item[] \textbf{10. Inclusion of Alternative Communicative Modes} \\
    \textbf{Theoretical Origin:} Young (1996); Bächtiger et al. (2010) (Type II Deliberation). \\
    \textbf{Definition:} Captures non-canonical speech forms—such as narrative/storytelling, rhetoric, and formal greeting—that complement formal argumentation by bridging cultural, ideological, or experiential divides in parliamentary committees. \\
    \textbf{Coding Categories:}
    \begin{itemize}[leftmargin=*]
        \item[] \textbf{Narrative / Testimony:} Use of situated personal stories, constituent accounts, or lived experiences to demonstrate legislative impact or illuminate marginalized perspectives.
        \item[] \textbf{Rhetoric / Expressive Language:} Use of emotional resonance, figurative language, or metaphor to contextualize policy claims and engage listeners.
        \item[] \textbf{Greeting and Deference:} Use of formal politeness, honorifics, or conciliatory gestures that acknowledge the standing of opponents and establish deliberative trust.
    \end{itemize}
\end{itemize}

\paragraph{Hybrid CB.}

\begin{itemize}[leftmargin=*]
    \item[] \textbf{Metric 1: Interactivity and Engagement} \\
    \textbf{Definition:} This metric measures the degree to which participants directly respond to the claims of others, rather than engaging in parallel monologues. In a healthy deliberation, participants listen, ask genuine questions to solicit information, and build upon previous speakers' points. Empirically, Knesset committees exhibit high levels of interactivity, but it frequently devolves into disruptive interruptions or cross-talk (e.g., shouting over one another during debates about terror and academia). Therefore, this metric evaluates both \textit{positive interactivity} (substantive responses and reciprocal dialogue) and \textit{negative interactivity} (disruptive interruptions that derail the speaker's thought process).

    \item[] \textbf{Metric 2: Level of Justification and Epistemic Reliance} \\
    \textbf{Definition:} This metric assesses whether participants support their demands with evidence, logic, and expert knowledge, rather than mere assertion. High-quality discourse includes ``sophisticated justifications'' where multiple complete inferences are made. Furthermore, in a systemic deliberative approach, good discourse relies on the appropriate integration of expert knowledge. In the Knesset transcripts, this is observed through the heavy reliance on legal advisors (e.g., from the Ministry of Justice or the committee's own counsel) to define legal boundaries, correct factual inaccuracies, and structure the consequences of legislation. The metric evaluates how well politicians incorporate this expert data into their arguments versus dismissing it for purely partisan reasons.

    \item[] \textbf{Metric 3: Civility, Respect, and Conflict Management} \\
    \textbf{Definition:} This evaluates the extent to which participants treat opposing groups, demands, and counterarguments with respect, which is a fundamental prerequisite for serious listening. In ideal deliberation, counterarguments are acknowledged and valued. The empirical Knesset data shows that committee meetings often struggle with this metric, featuring ad hominem attacks, sarcasm, and severe disrespect (e.g., calling opposing members a ``fifth column'' or comparing political opponents to terrorist organizations). This metric measures the frequency of degrading language versus respectful disagreement, tracking whether the discourse maintains a baseline of mutual recognition even during fierce ideological disputes.

    \item[] \textbf{Metric 4: Inclusiveness and Communicative Pluralism} \\
    \textbf{Definition:} This metric assesses whether diverse voices, particularly those of marginalized groups and non-politicians, are given an equal and fair opportunity to influence the debate. Drawing on ``Type II'' deliberative theory, it recognizes that rational, dispassionate argumentation can sometimes exclude the less privileged, meaning that rhetoric, emotion, and storytelling are valid and necessary forms of democratic communication. Empirically, this is vital in the Knesset, where bereaved parents use highly emotional, personal narratives to urge legislative action, and minority Arab MKs express the lived reality of their constituencies regarding soaring crime rates. This metric evaluates how well the committee makes space for, and seriously considers, these diverse perspectives and expressive styles.

    \item[] \textbf{Metric 5: Constructive Politics and Problem-Solving} \\
    \textbf{Definition:} This metric measures the committee's ability to move beyond entrenched, positional politics toward generating alternative or mediating proposals that aim for consensus or practical solutions. True deliberation in a legislative setting must ultimately aim to reach a binding, workable decision. The Knesset transcripts frequently display this metric in action: despite intense partisan arguments, members, legal advisors, and government representatives engage in real-time bargaining to tweak the wording of a law, adjust implementation timelines, or compromise on budget allocations. The metric captures the density of these constructive, solution-oriented exchanges.

    \item[] \textbf{Metric 6: Moderator (Chairperson) Efficacy} \\
    \textbf{Definition:} This metric evaluates the behavior and effectiveness of the committee chairperson in facilitating the deliberative process. A good moderator acts as a ``promoter of deliberation'' by managing the agenda, ensuring marginalized voices are heard, bringing the discussion back on topic, and intervening in conflicts. In the Knesset, the Chairperson (HaYo"r - originally in Hebrew letters)
     plays a highly active role in maintaining order, threatening to remove disruptive members, pushing for compromise, and clarifying procedural next steps. This metric tracks the chair's success in balancing the enforcement of rules with the allowance of free and open debate.

    \item[] \textbf{Metric 7: Common Good vs. Sectorial Framing} \\
    \textbf{Definition:} This metric evaluates the content of the justifications used by participants, specifically whether they appeal to narrow group interests or to the broader common good. Appeals to the common good can be utilitarian (the greatest good for the greatest number) or based on the difference principle (helping the least advantaged). In the empirical data, we see constant tension between sectorial lobbies (e.g., the agricultural lobby fighting to protect local honey production, or specific municipal authorities demanding resources) and the overarching national interest (e.g., lowering the cost of living or ensuring national security). This metric tracks how frequently and successfully participants link their specific, narrow claims to generalizable principles that benefit the entire society.
\end{itemize}

\section{Proxy Judge Model Selection}
\label{sec:appendix_judge_selection}

\paragraph{Human benchmark.} Two annotators independently rated 162 three-turn exchanges as high (1) or low (0) deliberative quality; scoring an inter-annotator agreement of 0.79 Krippendorff’s $\alpha$. The benchmark consists of the 148 exchanges on which both annotators agreed (40 positive, 108 negative); the 14 disagreements were excluded. The judge received the same instrument as the annotators: the coder definitions and clarifications, the session metadata, and the context turns surrounding the exchange. The clarifications were refined on a separate set of 21 expert-reviewed exchanges, none of which appear in the benchmark.

\paragraph{Model selection.} We evaluated seven OpenAI model settings (GPT-5.6-luna,
GPT-6-luna, GPT-6-luna-pro (reasoning effort set to `pro'), GPT-5.6-sol, GPT-6-sol, GPT-6.1-sol and GPT-6-astra), none of which were used in the CBs' reliability evaluation. The benchmark was split, stratified by label, into a validation set of 98 exchanges (26 positive) and a held-out test set of 50 exchanges (14 positive). We ran a stratified 5-fold cross-validation and selected the model with the highest mean macro F1. Validation runs were issued through the OpenRouter chat API on October 12, 2026; after validation, the selected model scored the full corpus through the OpenRouter Batch API with the identical prompt. On the held-out test set, GPT-6-luna achieved an accuracy of 82\% and a macro F1 of 77\%, against the human labels.

\section{Proxy C-DQI Distributions}
\label{sec:appendix_proxy_dist}

\begin{figure*}[ht]
  \centering
  \includegraphics[width=\linewidth]{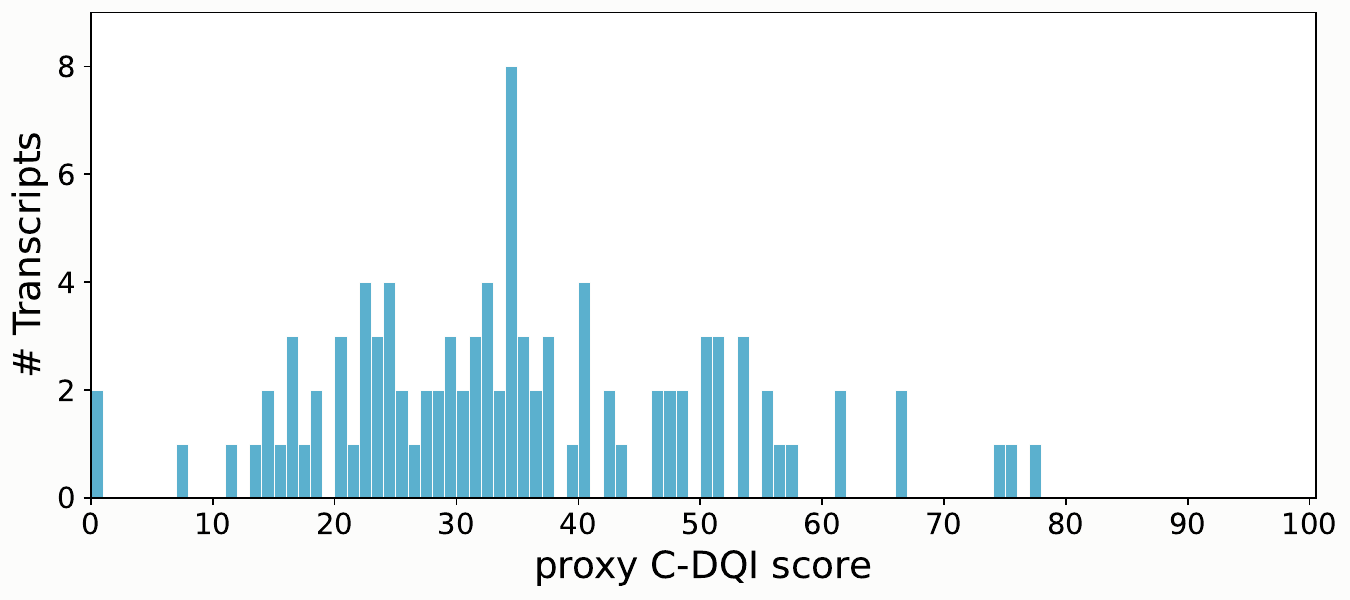}
  \caption{Distribution of the proxy C-DQI across the 100 evaluation transcripts, computed by averaging each transcript’s 3-turn exchange scores (36,341 in total), judged by GPT-6-luna.}
  \label{fig:cdqi-dist-luna}
\end{figure*}

In \S\ref{sec:results_acc}, we evaluate the validity of the generated CBs by correlating their scores with an extrinsic proxy measure of discourse quality. To ensure this proxy serves as a valid and discriminatory baseline, we analyzed the distribution of its scores across the 100 evaluation transcripts.

\autoref{fig:cdqi-dist-luna} illustrates the distribution of the proxy C-DQI scores as evaluated by GPT-6-luna. The scores exhibit a wide variance across the corpus, confirming that the proxy captures a broad spectrum of deliberative quality rather than collapsing into a narrow, uniform range. This distribution underscores the structural and topical diversity of the selected Knesset committee transcripts (as highlighted in Section~\ref{sec:data}) and provides a robust numerical foundation for the rank correlation analyses presented in Table~\ref{tab:proxy_accuracy}.

\section{Per-Model Correlation with the Proxy}
\label{sec:appendix_b}

Table~\ref{tab:model_codebook_proxy} reports Spearman's $\rho$ between each CB's discussion-level score and the proxy judge (GPT-6-luna), broken out per annotator model rather than averaged. The top-down CB aligns stronger with the proxies, for four out of the six models.
The top-Down and Vanilla DQI CBs exhibit the widest spread across models ($\Delta=0.14$); whereas the Bottom-Up CBs are the most stable ($\Delta \in [0.09,0.1]$).

\begin{table*}[t]
\centering
\resizebox{\textwidth}{!}{%
\begin{tabular}{lrrrrrrr}
\toprule
CB & Claude & Gemma4 & DictaLM & Qwen3.6 & GPT-OSS & Mistral \\
\midrule
Bottom-Up-Corrected & 0.57 & 0.55 & \textbf{0.64} & 0.59 & 0.58 & 0.58 \\
Bottom-Up-English   & \textbf{0.59} & 0.56 & 0.63 & 0.54 & 0.58 & \textbf{0.64} \\
Bottom-Up-Hebrew    & 0.57 & 0.53 & 0.54 & 0.55 & 0.53 & 0.62 \\
Top-Down            & 0.55 & \textbf{0.64} & 0.59 & \textbf{0.69} & \textbf{0.65} & \textbf{0.64} \\
Hybrid              & 0.54 & 0.56 & 0.52 & 0.61 & 0.59 & 0.63 \\
Vanilla DQI         & 0.48 & 0.53 & 0.41 & 0.55 & 0.49 & 0.53 \\
\bottomrule
\end{tabular}%
}
\caption{Spearman's $\rho$ between each C-DQI CB's discussion-level score and the proxy (GPT-6-luna), broken out by annotator model rather than averaged across models. All correlations significant at $p < 10^{-4}$. }
\label{tab:model_codebook_proxy}
\end{table*}

\end{document}